\documentclass[a4paper,conference]{IEEEtran}
\usepackage{xcolor}
\usepackage{amssymb}
\usepackage{booktabs}
\usepackage{siunitx}
\usepackage[pdftex]{graphicx}
\usepackage{verbatim}
\usepackage{hyperref}
\usepackage{tabularx,booktabs}
\usepackage{multirow}
\usepackage{lipsum}
\usepackage{array}
\usepackage{color, colortbl}
\usepackage[T1]{fontenc}
\usepackage{tikz}

\usepackage{cite}

\ifCLASSINFOpdf
\else
\fi
\usepackage{amsmath}

\newcommand{\mcl}[1]{\ensuremath{\mathcal{#1}}}

\newcommand{\vect}[1]{\ensuremath{\textbf{#1}}}

\newcolumntype{C}{>{\centering\arraybackslash}X}
\newcolumntype{L}{>{\centering\arraybackslash}p{0.2\textwidth}}
\newcommand{\NA}{---}

\let\svthefootnote\thefootnote
\newcommand\freefootnote[1]{%
  \let\thefootnote\relax%
  \footnotetext{#1}%
  \let\thefootnote\svthefootnote%
}

\begin{document}
%

\title{CARRADA Dataset: Camera and Automotive Radar with Range-Angle-Doppler Annotations}

%
\author{\IEEEauthorblockN{Arthur Ouaknine\IEEEauthorrefmark{1}\IEEEauthorrefmark{2},
Alasdair Newson\IEEEauthorrefmark{1},
Julien Rebut\IEEEauthorrefmark{2},
Florence Tupin\IEEEauthorrefmark{1} and
Patrick P\'erez\IEEEauthorrefmark{2}}
\IEEEauthorblockA{\IEEEauthorrefmark{1}LTCI, T\'el\'ecom Paris, Institut Polytechnique de Paris, 19 Place Marguerite Perey, 91120 Palaiseau, France}
\IEEEauthorblockA{\IEEEauthorrefmark{2}valeo.ai, 15 rue de la Baume, 75008 Paris, France\\ Email: arthur.ouaknine@telecom-paris.fr}}


\maketitle

\begin{abstract}
High quality perception is essential for autonomous driving (AD) systems. 
To reach the 
accuracy and robustness that are required by such systems, several types of sensors must be combined.  
Currently, mostly cameras and laser scanners (lidar) 
are deployed to build a representation of the world around the vehicle. 
While radar 
sensors have been used for a long time in the automotive industry, they are still under-used for AD 
despite their appealing characteristics (notably, their ability to measure the relative speed of obstacles and to operate even in adverse weather conditions).
To a large extent, this situation is due to the relative lack of automotive datasets with real radar signals that are both raw and annotated. 
%
In this work, we introduce CARRADA, a dataset of 
synchronized camera and radar recordings with range-angle-Doppler annotations. 
We also present a semi-automatic annotation approach, which was used to annotate the dataset, 
and a radar semantic segmentation baseline, which we evaluate on several metrics.
Both our code and dataset are available online.\footnote{\url{https://github.com/valeoai/carrada\_dataset}}
\end{abstract}



%
\IEEEpeerreviewmaketitle

\freefootnote{
\textcopyright 2021 IEEE. Personal use of this material is permitted. Permission from IEEE must be obtained for all other uses, in any current or future media, including reprinting/republishing this material for advertising or promotional purposes, creating new collective works, for resale or redistribution to servers or lists, or reuse of any copyrighted component of this work in other works. DOI: \href{https://doi.org/10.1109/ICPR48806.2021.9413181}{10.1109/ICPR48806.2021.9413181}}

\section{Introduction}

Advanced driving assistance systems (ADAS) and autonomous driving require a detailed understanding of complex driving scenes. 
With safety as the main objective, complementary and redundant sensors are mobilized to tackle this challenge. 
%
The best current systems rely on deep learning models that are trained on large annotated datasets for tasks such as object detection or semantic segmentation in video images and lidar point clouds. 
%
To improve further the performance of AD systems, extending the size and scope of open annotated datasets is a key challenge.



ADAS and AD-enabled vehicles are usually equipped with cameras, lidars and radars to gather complementary information from their environment and, thus, to allow as good a scene understanding as possible in all situations. 
Unfortunately, bad weather conditions are challenging for most of sensors: 
lidars have poor robustness to fog \cite{bijelic_benchmark_2018}, rain or snow; cameras behave poorly in low lighting conditions or in case of sun glare. Radar sensors, on the other hand, generate electromagnetic wave signals that are not affected by weather conditions or darkness. Also, radar 
informs not only about the 3D position of other objects, as lidar, but also about their relative speed (radial velocity).   
%
However, in comparison to other sensory data, radar signals are difficult to interpret, very noisy and with a low angular resolution. 
This is one reason why cameras and lidars have been preferred for the past years. 

\begin{figure}[!t]
\centering
\includegraphics[width=0.44\textwidth]{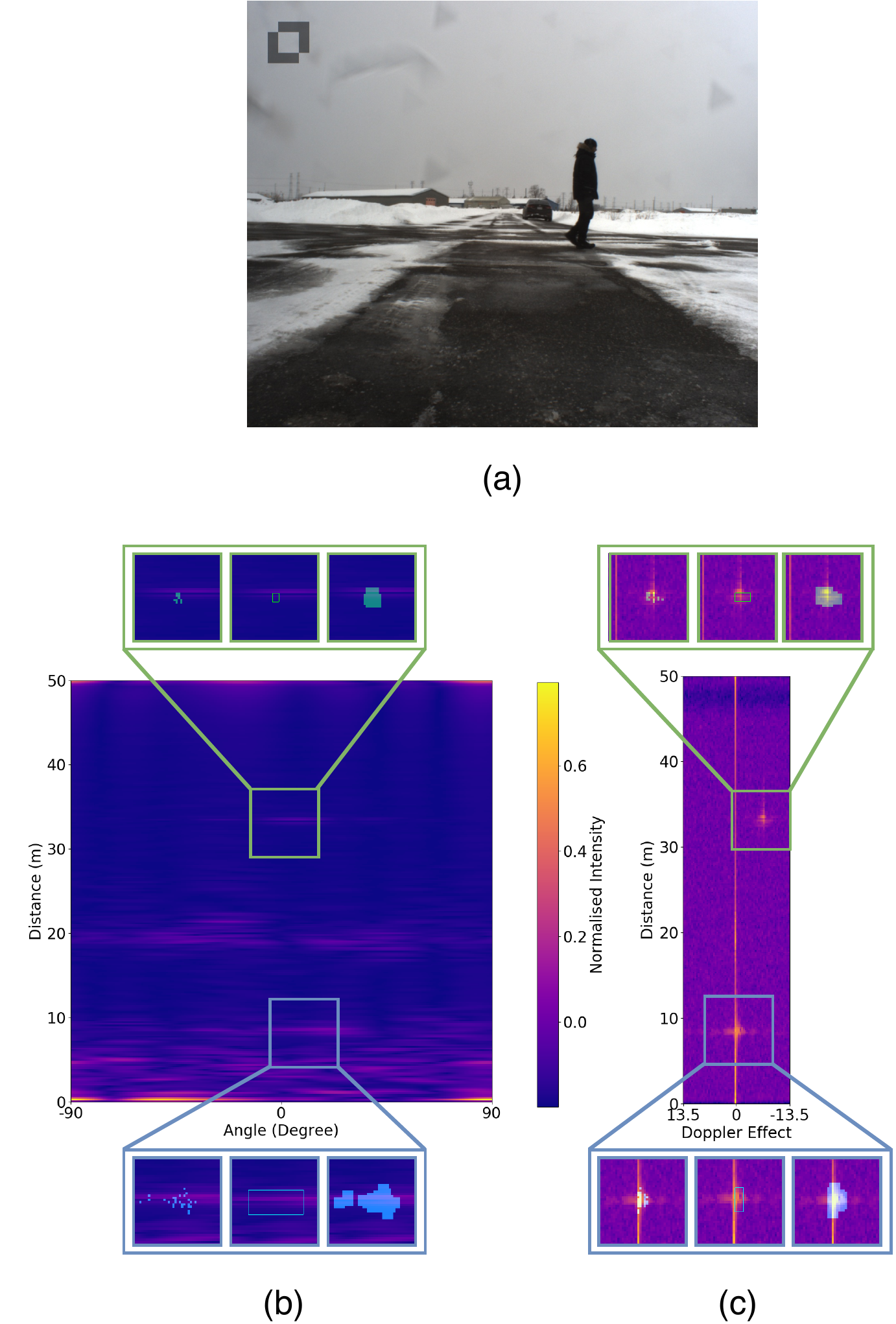}
\caption{\textbf{A scene from CARRADA dataset, with a pedestrian and a car.} (a) Video frame provided by the frontal camera, showing a pedestrian at approximately 8m from the sensors and a car in the background at approximately 33m; 
(b-c) Radar signal at the same instant in range-angle and range-Doppler representation respectively. Three types of annotations are provided: sparse points, bounding boxes and dense masks. The blue squares correspond to the pedestrian and the green ones to the car.
}
\label{fig_one_frame_labelled}
\end{figure}

In this paper, we present two main contributions. 
Firstly, we introduce 
CARRADA, a dataset with synchronized camera data and raw radar sequences together with range-angle-Doppler annotations for scene understanding (see a sample in Fig.\,\ref{fig_one_frame_labelled}). Annotations with bounding boxes, sparse points and dense masks are provided for range-Doppler and range-angle representations of the radar data. Each object has a unique identifier, being categorized as a \textit{pedestrian}, a \textit{car} or a \textit{cyclist}.
This dataset could be used for object detection, semantic segmentation (as illustrated in our segmentation baseline) or tracking in raw radar signals. It should also encourage sensor fusion in temporal data. 
Secondly, we describe a semi-automatic method to generate the radar annotations using only the camera information instead of a lidar as usual \cite{weston_probably_2019, lim_radar_2019, major_vehicle_2019}. 
The aim of this contribution is to reduce annotation time and cost by exploiting visual information without the need for an expensive sensor. 
A baseline for radar semantic segmentation is also proposed and evaluated on well-known metrics. We hope that it will encourage deep learning research applied to raw radar representations.


The paper is organised as follows. In Section \ref{background}, we discuss the related work and provide background on radar signals. Section \ref{dataset} introduces the proposed dataset and its acquisition setup. 
Section \ref{annotations} explains our semi-automatic annotation method, from visual and physical information to the labelling of radar signals with temporal tracking. Section \ref{baseline} details a baseline for radar semantic segmentation on raw representations. Finally, we discuss the proposed dataset and its current limitations in Section \ref{discussions}, before concluding in Section \ref{conclusion}.


\section{Background}
\label{background}

\subsection{Related work}
\label{related_work}

Previous works have applied deep learning algorithms to range-Doppler radar representation. Indoor activity recognition \cite{kim_human_2016} and gait classification \cite{klarenbeek_multi-target_2017} have been explored. Privacy motivates this application, as cameras can thus be avoided for scene understanding. Hand gesture recognition has been an active field of research using millimeter-wave radar for classification \cite{wang_interacting_2016, kim_hand_2016, dekker_gesture_2017, zhang_latern_2018, zhang_u-deephand_2019} or object signature recognition \cite{sun_automatic_2019, wang_rammar_2019}. Sensor fusion using radar and cameras has been studied for hand gesture classification \cite{molchanov_multi-sensor_2015}. Outdoor applications have also been considered to classify models of Unmanned Aircraft Vehicles (UAV) \cite{brooks_temporal_2018}.

Driving applications have recently shown an interest for radar sensors,
using representations that depend on the target task:
Doppler spectrograms for vehicle classification \cite{capobianco_vehicle_2018}, range-angle for object classification \cite{patel_deep_2019} or odometry \cite{aldera_what_2019}, range-angle and range-Doppler for object detection \cite{major_vehicle_2019}. Radar data are also used for position prediction with range-Doppler \cite{zhang_object_2020} or object box detection in images \cite{nabati_rrpn_2019} with only a few data points. Sensor fusion is considered for driving applications such as occupancy grid segmentation \cite{lekic_automotive_2019} or object detection in range-angle \cite{lim_radar_2019} while using radar and camera.

Scene understanding for autonomous driving using deep learning requires a large amount of annotated data. This challenge is well known by the community and open-source datasets have emerged in the past few years, \textit{e.g.}, \cite{geiger_vision_2013, yu_bdd100k_2020, cordts_cityscapes_2016, huang_apolloscape_2020, sun_scalability_2020}. Several types of annotations are usually provided, notably 2D or 3D bounding boxes and semantic segmentation masks for each object. They describe video frames from camera or 3D point clouds from lidar sensor. None of these 
datasets provides raw radar data recordings synchronized with the other sensors. Only very recent datasets include radar signals, but they are usually pre-processed and barely annotated. 
Raw data have a higher level of noise but provide information about all reflective objects in the scene.

The nuScenes dataset \cite{caesar_nuscenes_2020} is the first large-scale dataset providing radar data alongside lidar and camera data. However, the radar data are released with a non-annotated processed representation with only tens of points per frame. 
The Oxford Radar Robot Car dataset \cite{barnes_oxford_2020} groups camera, lidar and radar data for odometry. Only raw radar data with a range-angle representation are available, and they are not annotated for scene understanding.
Astyx has released a small dataset with a camera, a lidar and a high definition radar \cite{meyer_automotive_2019}. Annotations with 3D bounding boxes are provided on each modality, using the lidar for calibration. Raw radar data are processed and provided as a high resolution point cloud, comparable to a lidar output with longer range. However, frames are limited to a few hundreds.
In \cite{gao_experiments_2019}, the authors describe a partially-annotated dataset with synchronised camera and raw radar data. A single object is recorded during each sequence. Bounding boxes are provided on both camera and range-angle representations with a calibration made by a lidar sensor.
Range-angle segmented radar data are provided by \cite{nowruzi_deep_2020} for occupancy grid map in Cartesian coordinates. The annotations are generated using odometry from scene reconstruction of camera images.

To the best of our knowledge, range-angle and range-Doppler raw radar data have not been previously released together, nor have the corresponding annotations for object detection, semantic segmentation and tracking been provided. Moreover, there is no related work of deep learning algorithms exploiting both range-angle and range-Doppler annotations at the same time. This dataset will encourage exploration of advanced neural-network architectures.

\subsection{Radar sensor and effects}

A radar sensor emits electromagnetic waves via one or several transmitter antennas (Tx). The waves are reflected by an object and received by the radar via one or several receiver antennas (Rx). The comparison between the transmitted and the received waveforms infers the distance, the radial velocity, the azimuth angle and the elevation of the reflector regarding the radar position \cite{ghaleb_micro-doppler_2009}. Most of the automotive radars use Multiple Input Multiple Output (MIMO) systems: each couple of Tx/Rx receives the reflected signal assigned to a specific Tx transmitting a waveform. 

Frequency-Modulated Continuous Wave (FMCW) radar transmits a signal, called a chirp \cite{brooker_understanding_2005}, whose frequency 
is linearly modulated over the sweeping period $T_s$: At time $t_s \in \{0, \cdots, T_s\}$, 
the emitted sinusoidal signal has frequency
\begin{equation} \label{eq:transmit_freq}
    f_s  = f_c + \frac{B}{T_s}t_s ,
\end{equation}
where $f_c$ is the carrier frequency and $B$ the bandwidth,
and its phase reads 
\begin{equation}
    \phi_E(t) = 2\pi f_s t.
\end{equation}

After reflection on an object at distance $r(t)$ from the emitter, the received signal has phase:
\begin{equation}
    \phi_R(t) = 2\pi f_s (t-\tau)  = \phi_E(t) - \phi (t),
\end{equation}
where $\tau=\frac{2 r(t)}{c}$ is the time delay of the signal round trip, with $c$ the velocity of the wave through the air considered as constant, and $\phi(t)$ is the phase shift:
\begin{equation}
    \phi (t) = 2 \pi f_s \tau = 2 \pi f_s \frac{2r(t)}{c}.
\end{equation}
Measuring this phase shift (or equivalently the time delay between the transmitted and the reflected signal) grants access to the distance between the sensor and the reflecting object.
%

Its radial velocity is accessed through the frequency shift between the two signals, a.k.a. the Doppler effect.
%
%
Indeed, the phase shift varies when the target is moving: 
\begin{equation}
    f_d = \frac{1}{2\pi} \frac{d \phi}{d t} = \frac{2 v_R}{c}f_s ,
\end{equation}
where $v_R = d r / d t$ is the radial velocity of the target object w.r.t. the radar.
This yields the frequency Doppler effect whereby frequency change rate between transmitted and received signals, $\frac{f_d}{f_s} = \frac{2 v_R}{c}$, depends linearly on the relative speed of the reflector. Measuring this Doppler effect therefore amounts to recovering the radial speed
%
\begin{equation}
    v_R = \frac{c f_d}{2 f_s}.
\end{equation}

The transmitted and received signals, $s_E$ and $s_R$ are compared with a mixer that generates a so-called Intermediate Frequency (IF) signal.
The transmitted signal term is filtered using a low-pass filter and digitized by an Analog-to-Digital Converter (ADC). In this manner, the recorded signal carries the Doppler frequencies and ranges of all reflectors.



Using the MIMO 
system with multiple Rx antennas, the time delay between the received signals of each Rx transmitted by a given Tx carries the orientation information of the object. Depending on the positioning of the antennas, the azimuth angle and the elevation of the object are respectively deduced from the horizontal and vertical pairs of Tx/Rx. The azimuth angle $\alpha$ is deduced from the variation between the phase shift of adjacent pairs of Rx. We have $\Delta \phi_\alpha = 2 \pi f_s \frac{2 h \sin(\alpha)}{c}$, where $h$ is the distance separating the adjacent receivers.

Consecutive filtered IF signals are 
stored in a frame buffer which is a time-domain 3D tensor: 
the first dimension corresponds to the chirp index; the second one is the chirp sampling defined by the linearly-modulated frequency range; the third tensor dimension indexes Tx/Rx antenna pairs.
 
The Fast Fourier Transform (FFT) algorithm applies a Discrete Fourier Transform (DFT) to the recorded data from the time domain to the frequency domain. 
The 3D tensor is processed using a 3D-FFT: a Range-FFT along the rows resolving the object range, a Doppler-FFT along the columns resolving the object radial velocity and an Angle-FFT along the depth resolving the angle between two objects.

The range, velocity and angle bins in the output tensor correspond to discretized values defined by the resolution of the radar. 
The range resolution is defined as  $\delta d = \frac{c}{2B}$. The radial velocity resolution $\delta v_R = \frac{c}{2 f_c T}$ is inversely proportional to the frame duration time. The angle resolution $\delta \alpha = \frac{c}{f_c N_{\text{Rx}} h \cos(\alpha)}$ is the minimum angle separation between two objects to be distinguished, with $N_{\text{Rx}}$ the number of Rx antennas and $\alpha$ the azimuth angle between the radar and an object at distance $D$ reflecting the signal.

The next section will describe the settings of the radar sensor used and recorded dataset.

\section{Dataset}
\label{dataset}

The dataset has been recorded in Canada on a test track to reduce environmental noise. The acquisition setup consists of an FMCW radar and a camera mounted on a stationary car. 
The radar uses the MIMO system configuration with 2 Tx and 4 Rx producing a total of 8 virtual antennas. The parameters and specifications of the sensor are provided in Table \ref{table_radar_specs}. The image data recorded by the camera and the radar data are synchronized to have the same frame rate in the dataset.
The sensors are also calibrated to have the same Cartesian coordinate system. The image resolution is $1238 \times 1028$ pixels. The range-Doppler and range-angle representations are respectively stored in 2D matrices of size $256 \times 64$ and $256 \times 256$.

\begin{table}
\renewcommand{\arraystretch}{1.0}
\caption{Parameters and settings of the radar sensor.}
\label{table_radar_specs}
\centering
\begin{tabular}{cc} \toprule
Parameter & Value\\ \midrule
Frequency & 77 Ghz\\
Sweep Bandwidth & 4 Ghz\\
Maximum Range & 50 m\\ 
FFT Range Resolution & 0.20 m\\
Maximum Radial Velocity & 13.43 m/s\\
FFT Radial Velocity Resolution & 0.42 m/s\\
Field of View & \ang{180}\\
FFT Angle Resolution & \ang{0.70}\\ 
Number of Chirps per Frame & 64\\
Number of Samples per Chirp & 256\\ \bottomrule
\end{tabular}
\end{table}

\begin{table}[b!]
\renewcommand{\arraystretch}{1.0}
\caption{Parameters of the dataset.}
\label{table_dataset}
\centering
\begin{tabular}{cc} \toprule
Parameter & Value\\ \midrule
Number of sequences & 30\\
Total number of instances & 78\\
Total number of frames & 12666 (21.1 min)\\ 
Maximum number of frames per sequence & 1017 (1.7 min)\\
Minimum number of frames per sequence & 157 (0.3 min)\\
Mean number of frames per sequence & 422 (0.7 min)\\
Total number of annotated frames with instance(s) & 7193 (12.0 min)\\ \bottomrule
\end{tabular}
\end{table}

\begin{figure}[t!]
\centering
\includegraphics[width=3.4in]{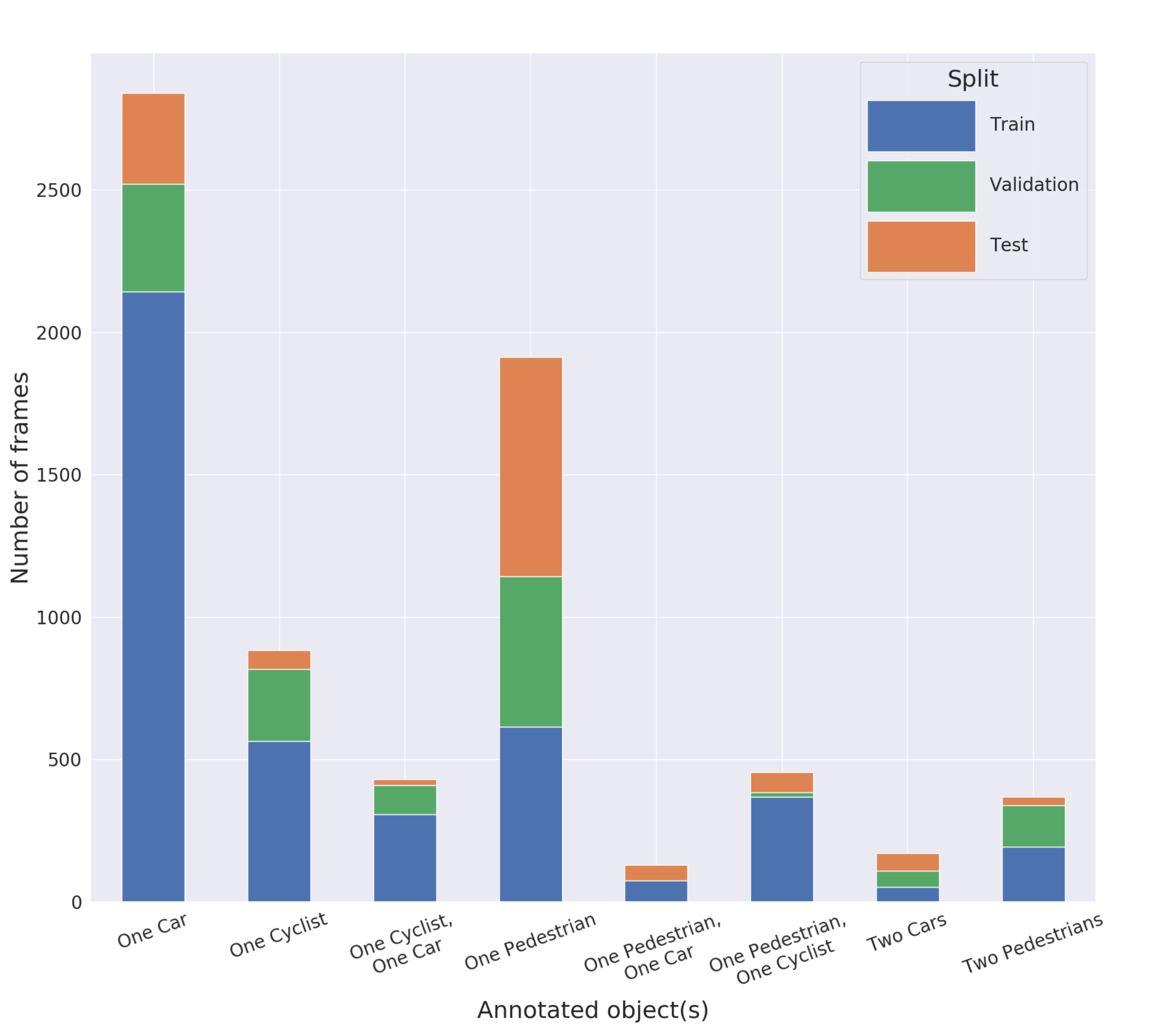}
\caption{\textbf{Object distribution across CARRADA.} Distribution of the eight object configurations present in the dataset, expressed as frame numbers across the three parts of the proposed split. 
}
\label{fig_frame_distrib}
\end{figure}

\begin{figure}[t]
\centering
\includegraphics[width=3.4in]{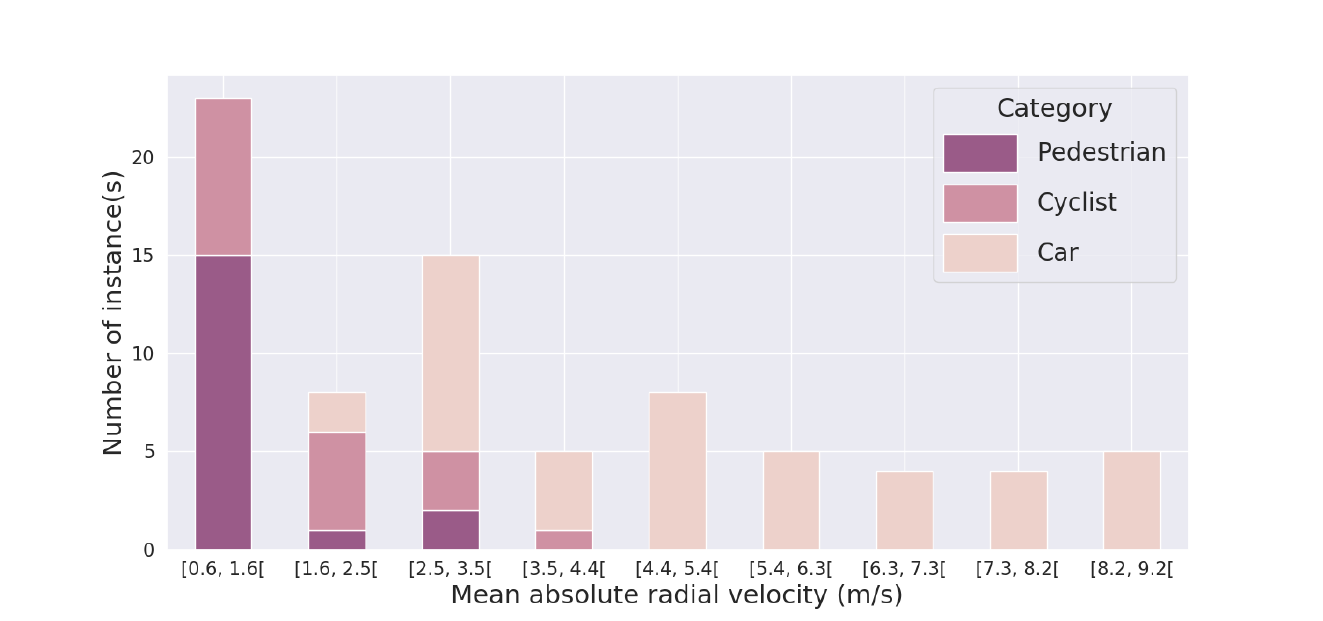}
\caption{\textbf{Distribution of radial velocities for all categories.} For each annotated instance, its absolute radial velocity in m/s is averaged, over its sparse annotations in each frame and over time, 
and one histogram is built for each object class. Note that annotated velocities are actually signed (negative when the object if moving away and positive when it approaches the radar).   
}
\label{fig_velocity_distrib}
\end{figure}

\begin{figure*}[htb]
\centering
\includegraphics[width=6in]{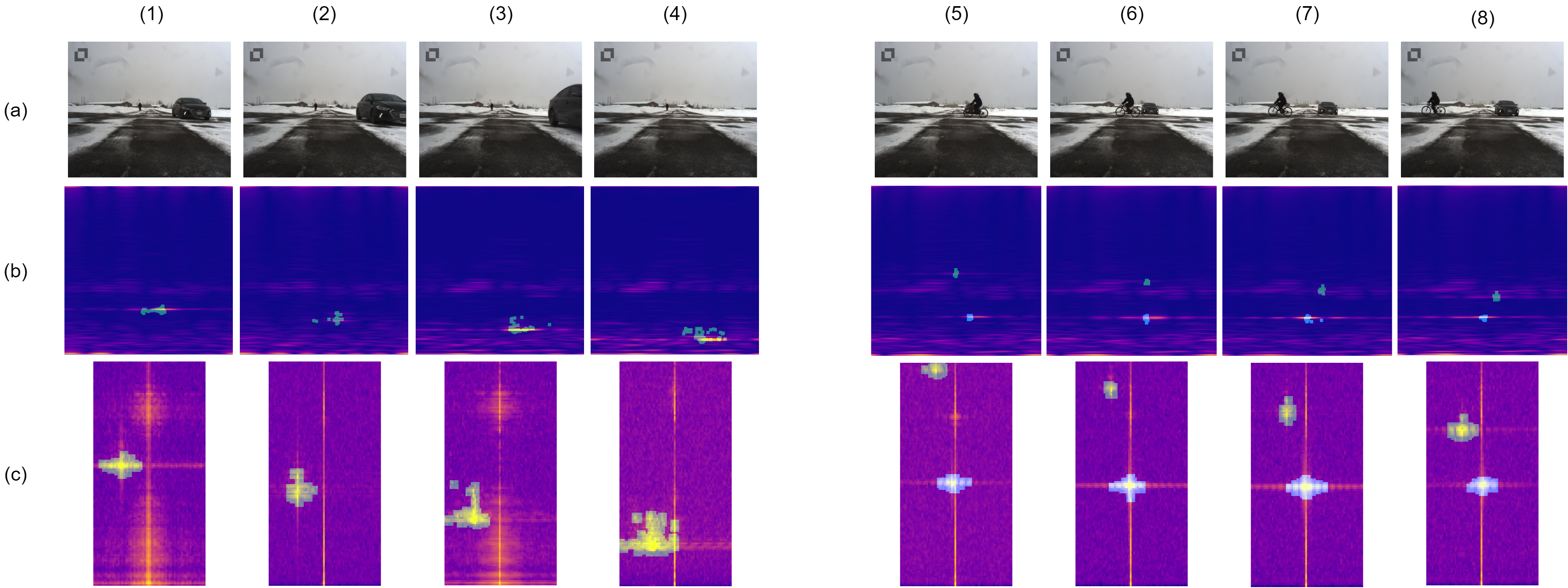}
\caption{\textbf{Two scenes from CARRADA dataset, one with a car, the other on with a cyclist and a car.}
(a) Video frames provided by the frontal camera showing moving objects in a fixed environment; (b-c) Radar signals at the same instants in range-angle and cropped range-Doppler representation  respectively. (1-4) First sequence; (5-8) Second sequence. Both sequences are selected with a timestamp interval of 5 frames. In the first sequence, the segmentation mask corresponds to the annotation of the approaching car in the scene. Range-Doppler data (c)(1) and (c)(3) show that our method is robust to recording noise. In frames (4), the car is still in the radar's  field of view but it has disappeared from the camera. In the second sequence, the segmentation masks correspond respectively to the annotations of the moving cyclist (blue) and car (green). The cyclist is moving from right to left in front of the radar, its radial velocity is progressively changing from positive to negative.}
\label{fig-temporal_example} 
\end{figure*}

Scenarios with cars, pedestrians and cyclists have been recorded. 
One or two objects are moving in the scene at the same time with various trajectories to simulate urban driving scenarios. 
The distribution of these scenarios across the dataset is shown in Figure \ref{fig_frame_distrib}.
The objects are moving in front of the sensors: approaching, moving away, going from right to left or from left to right (see examples in Figure \ref{fig-temporal_example}). Each object is an \textit{instance} tracked in the sequence. The distribution of mean radial velocities for each object category is provided in Figure \ref{fig_velocity_distrib}, while other global statistics about the recordings can be found in Table \ref{table_dataset}.

Object signatures are annotated in both range-angle and range-Doppler representations for each sequence. Each instance has an identification number, a category and a localization in the data. Three types of annotations for localization are provided: sparse points, boxes and dense masks. The next section will describe the pipeline used to generate them.



\section{Pipeline for annotation generation}
\label{annotations}

Automotive radar representations are difficult to understand compared to natural images. Objects are represented by shapes with varying sizes carrying physical measures. It is not a trivial task to produce good quality annotations on this data.
This section details a semi-automatic pipeline based on video frames to provide annotations on radar representations.

\subsection{From vision to physical measurements}
\label{from_vision}

The camera and radar recordings are synchronized. Visual information in the natural images is used to get physical prior knowledge about an instance as well as its category. The real-world coordinates of the instance and its radial velocity are estimated generating the annotation in the radar representation. This first step instantiates a tracking pipeline propagating the annotation in the entire radar sequence. 

Each video sequence is processed by a Mask R-CNN \cite{he_mask_2017} model providing both semantic segmentation and bounding box predictions for each detected instance. Both are required for our pipeline to compute the center of mass of the object and to track it. 
Instance tracking is performed with the Simple and Online Real time Tracking (SORT) algorithm \cite{bewley_simple_2016}. 
This light-weight tracker computes the overlap between the predicted boxes and the tracked boxes of each instance at the previous frame. The selected boxes are the most likely to contain the same instance, i.e. the boxes with the highest overlap.

The center of mass of each segmented instance is projected on the bottom-most pixel coordinates of the segmentation mask. This projected pixel localized on the ground is considered as the reference point of the instance.
Using the intrinsic and extrinsic parameters of the camera, pixel coordinates of a point in the real-world space are expressed as:
\begin{equation}
s \; \vect{p} = A \; B \; \vect{c} ,
\end{equation}
where $\vect{p} = [p_x,p_y,1]^{\top}$ and $\vect{c}= [c_x,c_y,c_z,1]^{\top}$ are respectively the pixel coordinates in the image and the real-world point coordinates, $s$ a scaling factor, and $A$ and $B$ are the intrinsic and extrinsic parameters of the camera defined as:
\begin{equation}
A=
\begin{bmatrix}
f_x & 0 & a_x \\
0 & f_y & a_y \\
0 & 0 & 1
\end{bmatrix},\,
B=
\begin{bmatrix}
r_{11} & r_{12} & r_{13} & m_1 \\
r_{21} & r_{22} & r_{23} & m_2 \\
r_{31} & r_{32} & r_{33} & m_3
\end{bmatrix}.
\end{equation}

Using this equation, one can determine $\vect{c}$ knowing $\vect{p}$ with a fixed value of elevation.

Regarding a given time interval $\delta t$ separating two frames $t-\delta_t$ and $t$, the velocity vector $\mathbf{v}^t$ is defined as:
\begin{equation}
    \mathbf{v}^t = \vect{c}^t - \vect{c}^{t-\delta t},
\end{equation}
where $\vect{c}^t$ is the real-world coordinate in frame $t$.
The time interval chosen in practice is $\delta t = 1 $ second.

\begin{figure}[!t]
\centering
\includegraphics[width=3in]{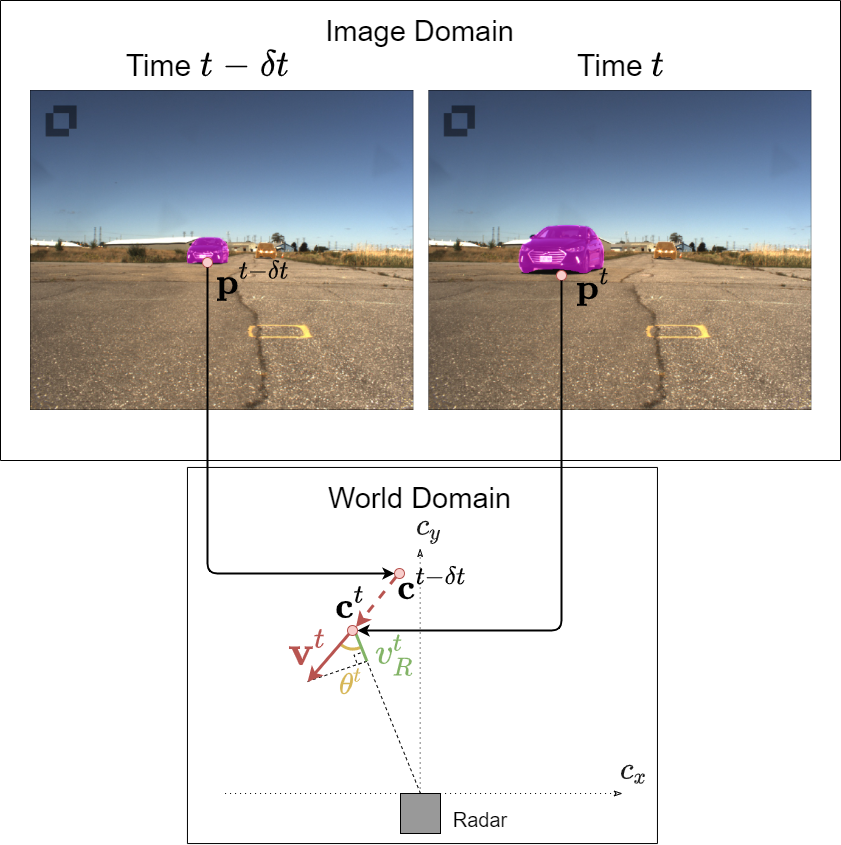}
\caption{\textbf{Estimation of the radial velocity from natural images.}
The space $(c_x, c_y, c_z)$ defines real-world coordinates regarding to the radar, $c_z$ is fixed to zero. Points in the real-world domain $\vect{c}^{t-\delta t}$ and $\vect{c}^{t}$ (bottom) are estimated using the points in the pixel domain $\vect{p}^{t-\delta t}$ and $\vect{p}^{t}$ (top). The velocity vector $\vect{v}^t$ is estimated with the real-world points. The radial velocity of the object at time $t$ corresponds to the projection of its velocity vector on the straight line between the radar of the object.}
\label{fig_relative_velocity}
\end{figure}

The Doppler effect recorded by the radar is equivalent to radial velocity of the instance reflecting the signal. The radial velocity $v_R^t$ at a given frame $t$ is defined as:
\begin{equation}
    v_R^t = \cos{\theta^t} \; \lVert \mathbf{v}^t \rVert ,
\end{equation}
where $\theta^t$ is the angle formed by $\mathbf{v}^t$ and the straight line between the radar and the instance. The quantization of the radial velocity is illustrated in Figure \ref{fig_relative_velocity}.
This way, each instance detected in the frame is characterized by a feature point $I^t = [\vect{c}^t, v_R^t]^{\top}$.
This point will be projected in a radar representation to annotate the raw data and track it in this representation.

\subsection{DoA clustering and centroid tracking}

\begin{figure}[!t]
\centering
\includegraphics[width=3in]{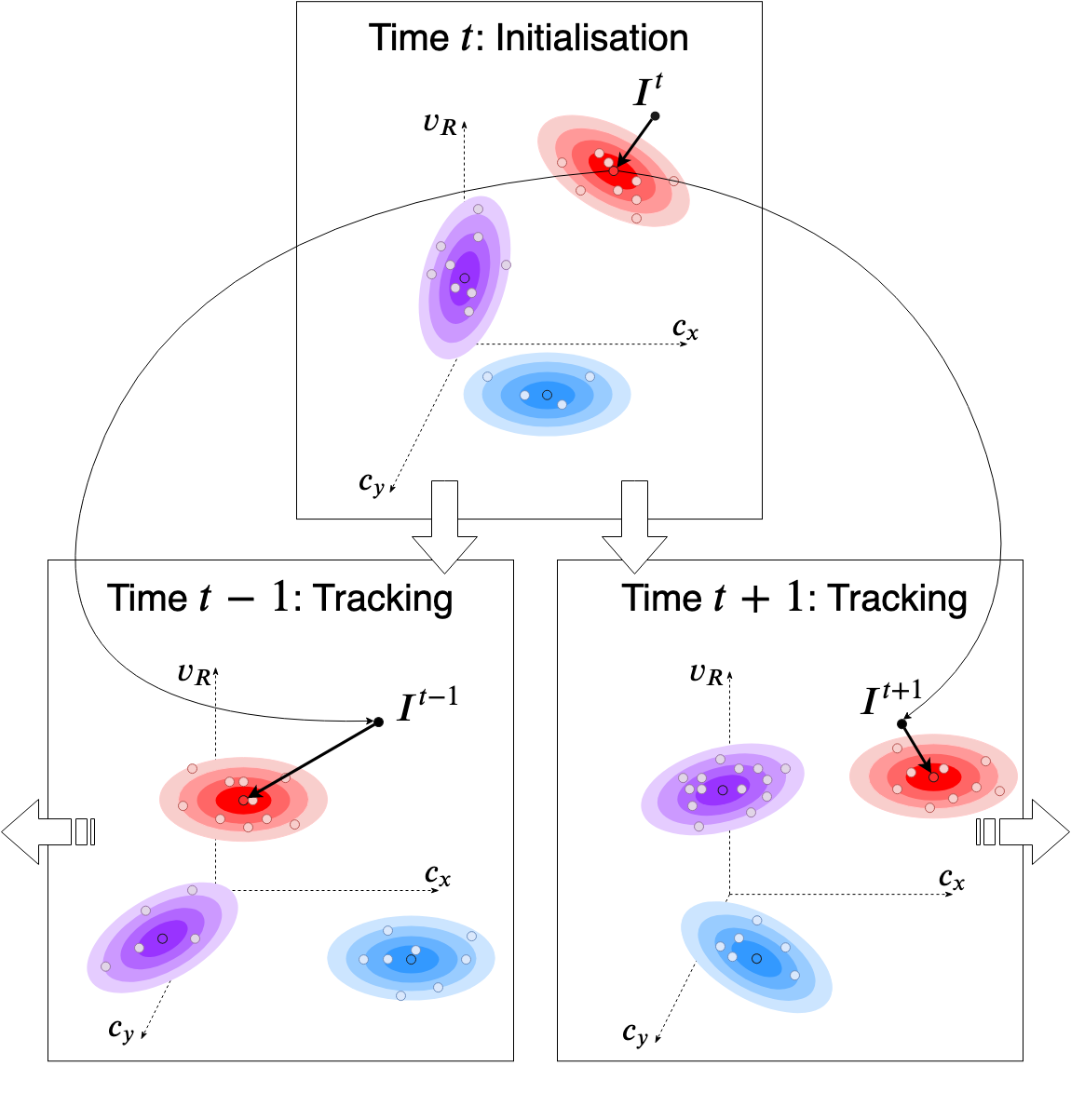}
\caption{\textbf{Tracking of the Mean Shift cluster to propagate the annotation in the sequence.}
The Mean Shift algorithm used with the bandwidth selection method is applied to the DoA-Doppler representation at time $t$. The estimated point $I^t$, using the computer vision pipeline and corresponding to the tracked object, is associated to its closest cluster. The centroid of this cluster considered as $I^{t+1}$ and $I^{t-1}$ in the next and previous DoA-Doppler frames is tracked iteratively.}
\label{fig_clustering_tracking}
\end{figure}

The range-angle representation is a radar scene in polar coordinates. Its transformation into Cartesian coordinates is called Direction of Arrival (DoA). Points are filtered by a Constant False Alarm Rate (CFAR) algorithm \cite{rohling_radar_1983} keeping the highest intensity values while taking into account the local relation between points. The DoA is then a sparse point cloud in a 2D coordinate space similar to a Bird's Eye View (BEV) representation.
The representation is enhanced using the recorded Doppler for each point. The 3D point cloud combines the Cartesian coordinates of the reflected point and its Doppler value. This helps to distinguish the signature boundaries of different objects. The feature point $I^t$ is projected in this space and assigned to a cluster of points considered as the reflection of the targeted instance. It is then tracked in the past and future using the following process, illustrated in Figure \ref{fig_clustering_tracking}.

At a given timestamp chosen by the user, a 3D DoA-Doppler point cloud is clustered using the Mean Shift algorithm \cite{comaniciu_mean_2002}. Let $\{\vect{x}_0, \cdots, \vect{x}_{n-1} \}$ be a point cloud of $n$ points. For a given starting point, 
the algorithm iteratively computes a weighted mean of the current local neighborhood and updates the point until convergence. Each iteration reads:
%
\begin{equation}
    \vect{x} \leftarrow \frac{\sum_{i=0}^{n-1}\vect{x}_i K(\vect{x};\vect{x}_i, \sigma)}{\sum_{i=0}^{n-1}K(\vect{x};\vect{x}_i, \sigma)},
\end{equation}
where $K( \vect{x};\vect{x}_i,\sigma) = \mathcal{N}(\left\| \frac{\vect{x} - \vect{x}_i}{\sigma} \right\|;0,1)$ is the multivariate spherical Gaussian kernel with bandwidth $\sigma$, centered at $\vect{x}_i$.
All initial points leading to close final locations at convergence are considered as belonging to the same cluster.


Mean-Shift clustering is sensitive to the bandwidth parameter. Its value should depend on the point cloud distribution and it is usually defined with prior knowledge about the data. 
In our application, it is not straightforward to group points belonging to the same object in the DoA-Doppler point cloud representation. The number of points and their distribution depend on the distance and the surface of reflectivity of the target. Moreover, these characteristics change during a sequence while the instance is moving in front of the radar. Inspired by \cite{bugeau_bandwidth_2007}, an optimal bandwidth is automatically selected for each instance contained in each point cloud.


For a given DoA-Doppler point cloud, the closest cluster to the feature point $I^t$ is associated to an instance. Let $\sigma_b \in \{ \sigma_0, \cdots, \sigma_{B-1} \}$ be a bandwidth in a range of $B$ ordered values. A Mean-Shift algorithm noted $\text{MeanShift}(\sigma_b)$ selects the closest cluster $\mathcal{C}_b$ to $I^t$ containing $n_b$ points. After running the algorithm with all bandwidth values, $\{\mathcal{C}_0, \cdots, \mathcal{C}_{B-1} \}$ optimal clusters are found. The optimal bandwidth is selected by comparing the stability of the probability distribution of the points between the selected clusters.

For each $b \in \{0, \cdots, B-1\}$, the probability distribution $p_b$ estimated with the $n_b$ points of the cluster $\mathcal{C}_b =  \{\vect{x}_0, \cdots, \vect{x}_{n_b-1} \}$ is the Gaussian distribution $\mcl{N}(\widehat{\mu}_b, \widehat{\Sigma}_b)$  with expectation $\widehat{\mu}_b = \frac{1}{n_b} \sum_{i=0}^{n_b-1} \vect{x}_i$ and variance $\widehat{\Sigma}_b = \frac{1}{n_b-1} \sum_{i=0}^{n_b-1} (\mathbf{x}_{i \cdot} - \widehat{\mu}_b)(\mathbf{x}_{i \cdot} - \widehat{\mu}_b)^{\top}$.

%
%
Using these fitted distributions, the bandwidth $\sigma_{b^*}$ is selected by choosing the one which is the most ``stable'' with respect to a varying bandwidth: 
\begin{equation}
\begin{split}
    b^* = \underset{b \in \{1, \cdots, B-2\}}{\text{argmin}} & \Big[\textit{JS} \big( p_b \|  p_{b-1} \big)
     + \textit{JS} \big(p_b \| p_{b+1} \big)\Big],
\end{split}
\label{eq:jsMin}
\end{equation}
where $\textit{JS}$ is the Jensen-Shannon divergence \cite{endres_new_2003}. This is a proper metric derived from Kullback-Leibler ($\textit{KL}$) divergence \cite{kullback_information_1951} as 
$\textit{JS}(p\|q)^2 = \frac{\textit{KL}(p \| \frac{p + q}{2}) + \textit{KL}(q \| \frac{p + q}{2})}{2}$, for two probability distributions $p$ and $q$.



Once $\sigma_b$ is found, the closest cluster to $I^t$ using $\text{MeanShift}(\sigma_b)$ is considered as belonging to the targeted instance. The points $I^{t+1}$ and  $I^{t-1}$ are set with the centroid of this cluster. The process is then iterated in the previous and next frames to track  the center of the initial cluster until the end of the sequence.

\subsection{Projections and annotations}

We recall that $\mathcal{C}_{b^*}$ is the cluster associated to the point $I^t$ at time $t$ using $\text{MeanShift}(\sigma_{b^*})$, where $\sigma_{b^*}$ is the estimated optimal bandwidth. This cluster is considered as belonging to the tracked object. A category is associated to it by using the segmentation model on the image (Section \ref{from_vision}).
The points are projected onto the range-Doppler representation using the radial velocity and the distance is computed with the real-world coordinates. They are also projected onto the range-angle representation by converting the Cartesian coordinates to polar coordinates. 

Let $f_{D}$ be the function which projects a point from the DoA-Doppler representation into the range-Doppler representation. Similarly, we denote with $f_{A}$ the projection into the range-angle representation. The sets of points $M_D = f_D(\mathcal{C}_b)$ and $M_A = f_A(\mathcal{C}_b)$ correspond, respectively, to the range-Doppler and range-angle representations of $\mathcal{C}_b$. They are called the sparse-point annotations.

The bounding box of a set of points in $\mathbb{R}^2$ (either from $M_D$ or $M_A$) is defined as a rectangle parameterized by $\{ (x_{\min}, y_{\min}),  (x_{\max}, y_{\max})\}$ where $x_{\min}$ is the minimum $x$-coordinate of the set, $x_{\max}$ is the maximum, and similarly for the $y$-coordinates.

Finally, the dense mask annotation is obtained by dilating the sparse annotated set with a circular structuring element: Given the sparse set of points $ (x_0,y_0), \dots, (x_{N-1},y_{N-1})$, the associated dense mask is the set of discrete coordinates in $\cup_{i=0}^{N-1} \mcl{B}_r(x_i, y_i)$, where $\mcl{B}_r(x, y)$ is the disk of radius $r$ centered at $(x,y)$. 


In the following section, we propose a baseline for radar semantic segmentation trained and evaluated on the annotations detailed above.

\section{Baseline}
\label{baseline}
\begin{table*}[t]
 \caption{\textbf{Semantic segmentation performances (\%) on the test dataset for range-Doppler (RD) and range-angle (RA) representations}. Models are trained on dense mask annotations and evaluated on both dense mask (top values) and sparse points (bottom values in parenthesis) annotations. Results are evaluated with Intersection over Union (IoU), Pixel Precision (PP) and Pixel Recall (PR). Metrics are computed by category and aggregated with both arithmetical (m) and harmonic (h) means. Lines (\NA) replacing values indicate non-applicable metrics, for example IoU results on sparse annotations.}
\label{table_baseline}
\def\arraystretch{1.2}
\setlength\tabcolsep{3pt}
\scriptsize
\begin{tabularx}{\textwidth}{c @{\hskip 0.1in} c @{} CCCCCC CCCCCC CCCCCC @{}}
\toprule
 \multicolumn{1}{c}{Data} & \multicolumn{1}{c}{Model} & \multicolumn{6}{c}{IoU} &   \multicolumn{6}{c}{PP} & \multicolumn{6}{c}{PR} \\
 \cmidrule(lr{3pt}){3-8}
 \cmidrule(l{3pt}r{3pt}){9-14}
 \cmidrule(l{3pt}r){15-20}
 & & \rotatebox{90}{Background} & \rotatebox{90}{Pedestrian} & \rotatebox{90}{Cyclist} & \rotatebox{90}{Car} & \rotatebox{90}{mIoU} & \rotatebox{90}{hIoU}
 & \rotatebox{90}{Background} & \rotatebox{90}{Pedestrian} & \rotatebox{90}{Cyclist} & \rotatebox{90}{Car} & \rotatebox{90}{mPP} & \rotatebox{90}{hPP}
 & \rotatebox{90}{Background} & \rotatebox{90}{Pedestrian} & \rotatebox{90}{Cyclist} & \rotatebox{90}{Car} & \rotatebox{90}{mPR} & \rotatebox{90}{hPR} \\ 
\midrule
    \multirow{5}{*}{\textbf{RD}} 
    
    & \multirow{2}{*}{FCN-32s} & 99.6 (\NA)  &  16.8 (\NA)  &  3.2 (\NA)  &  27.5 (\NA)  &  36.8 (\NA)  &  8.1 (\NA)  &  99.6 (\NA)  &  69.4 \textbf{(17.3)}  &  5.4 (1.0)  &  64.2 \textbf{(17.2)}  &  59.7 (11.9)  &  13.8 (2.2)  &  99.9 (\NA)  &  20.3 (28.3)  &  6.7 (8.1)  &  32.3 (47.5)  &  39.8 (28.0)  &  13.3 (14.0)  \\
    
    & \multirow{2}{*}{FCN-16s} &  99.6 (\NA)  &  28.9 (\NA)  &  7.2 (\NA)  &  42.1 (\NA)  &  44.5 (\NA)  &  17.2 (\NA)  &  99.7 (\NA)  &  64.7 (15.1)  &  18.0 (4.7)  &  67.6 (17.0)  &  62.5 (12.3)  &  40.3 (8.6)  &  99.9 (\NA)  &  39.2 (50.9)  &  10.7 (16.9)  &  54.1 (74.1)  &  51.0 (47.3)  &  23.4 (27.5)  \\
    
    & \multirow{2}{*}{FCN-8s} &  99.7 (\NA)  &  \textbf{45.2} (\NA)  &  \textbf{15.5} (\NA)  &  \textbf{51.3} (\NA)  &  \textbf{52.9} (\NA)  &  \textbf{34.2} (\NA)  &  99.8 (\NA)  &  \textbf{72.3} (15.5)  &  \textbf{35.2} \textbf{(9.6)}  &  \textbf{69.8} (17.0)  &  \textbf{69.3} \textbf{(14.0)}  &  \textbf{59.8} \textbf{(13.1)}  &  99.9 (\NA)  &  \textbf{55.0} \textbf{(76.4)}  &  \textbf{22.1} \textbf{(35.7)}  &  \textbf{66.8} \textbf{(88.9)}  &  \textbf{60.9} \textbf{(67.0)}  &  \textbf{44.3} \textbf{(56.7)} \\

\midrule

\multirow{5}{*}{\textbf{RA}} 
    
    & \multirow{2}{*}{FCN-32s} &  99.8 (\NA) & 0.0 (\NA)  &  0.0 (\NA)  &  14.2 (\NA)  &  28.5 (\NA)  &  0.0 (\NA)  &  99.9 (\NA)  &  18.8 (6.1)  &  0.0 (0.0)  &  \textbf{69.3} \textbf{(13.7)}  &  47.0 (6.6)  &  0.0 (0.0)  &  100.0 (\NA)  &  0.0 (0.1)  &  0.0 (0.0)  &  15.5 (24.6)  &  28.9 (8.2)  &  0.0 (0.0)\\
    
    & \multirow{2}{*}{FCN-16s} &  99.8 (\NA)  &  0.9 (\NA)  &  0.0 (\NA)  &  13.7 (\NA)  &  28.6 (\NA)  &  0.0 (\NA)  &  99.9 (\NA)  &  39.8 (5.2)  &  \textbf{0.2} (0.0)  &  68.0 (12.2)  &  \textbf{52.0} (5.8)  &  \textbf{0.9} (0.0)  &  100.0 (\NA)  &  0.9 (1.7)  &  0.0 (0.0)  &  15.4 (22.7)  &  29.1 (8.1)  &  0.0 (0.0)\\
    
    & \multirow{2}{*}{FCN-8s} &  99.9 (\NA)  &  \textbf{5.5} (\NA)  &  0.0 (\NA)  &  \textbf{25.1} (\NA)  &  \textbf{32.6} (\NA)  &  \textbf{0.1} (\NA)  &  99.9 (\NA)  &  \textbf{42.2} \textbf{(10.0)}  &  0.1 (\textbf{0.1})  &  65.4 (12.4)  &  51.9 (\textbf{7.5})  &  0.6 (\textbf{0.2})  &  100.0 (\NA)  &  \textbf{6.3} \textbf{(11.3)}  &  0.0 (\textbf{0.1})  &  \textbf{30.0} \textbf{(45.5)}  &  \textbf{34.1} \textbf{(19.0)}  &  \textbf{0.1} \textbf{(0.3)} \\

\bottomrule

\end{tabularx}
\end{table*}

We propose a baseline for semantic segmentation using range-Doppler or range-angle radar representation to detect and classify annotated objects. Fully Convolutional Networks (FCNs) \cite{long_fully_2015} are used here to learn features at different scales
by processing the input data with convolutions and down-sampling. Feature maps from convolutional layers are up-sampled with up-convolutions to recover the original input size. Each bin of the output segmentation mask is then classified. The particularity of FCN is the use skip connections from features learnt at different levels of the network to generate the final output. We denote FCN-32s a network where the output mask is generated only by up-sampling and processing feature maps with $1/32$ resolution of the input. Similarly, FCN-16s is a network where $1/32$ and $1/16$ resolution features maps are used to generate the output mask. In the same manner, FCN-8s fuses $1/32$, $1/16$ and $1/8$ resolution feature maps for output prediction.

The models are trained to recover dense mask annotations with four categories: \textit{background}, \textit{pedestrian}, \textit{cyclist} and \textit{car}. The background corresponds to speckle noise, sensor noise and artefacts which are covering most of the raw radar data.
Parameters are optimized for 100 epochs using a categorical cross entropy loss function and the Adam optimizer \cite{kingma_adam_2015} with the recommended parameters ($\beta_1 = 0.9$, $\beta_2 = 0.999$ and $\epsilon = 1 \cdot 10^{-8}$). The batch size is fixed to 20 for the range-Doppler representation and to 10 for the range-angle representation to fill the memory capacities of the GPU. For both representations, the learning rate is initialized to $1 \cdot 10^{-4}$ for FCN-8s and $5 \cdot 10^{-5}$ for FCN-16s and FCN-32s. The learning rate has an exponential decay of 0.9 each 10 epochs. Training has been completed using the PyTorch framework with a single \textit{GeForce RTX 2080 Ti} GPU. 

Performances are evaluated for each radar representation using the Intersection over Union (IoU), the Pixel Precision (PP) and the Pixel Recall (PR) for each category. Metrics by category are aggregated using arithmetic and harmonic means.
To ensure consistency of the results, all performances are averaged from three trained models initialized with different seeds. 
Results are presented in Table \ref{table_baseline} \footnote{\textit{Erratum}: these results have been updated since the ICPR proceedings. Models are selected using the PP metric instead of the previously used PR.}. 

Models are trained on dense mask annotations and evaluated on both dense mask (top values) and sparse points (bottom values in parentheses) annotations. 
Sparse points are more accurate than dense masks, therefore evaluation on this type of annotation provides information on the behaviour of predictions on key points. 
However, localization should not be evaluated for sparse points using a model trained on dense masks, therefore IoU performances are not reported. 
The background category cannot be assessed for the sparse points because some of the points should belong to an object but are not annotated \textit{per se}. 
Thus, arithmetic and harmonic means of sparse points evaluations are computed for only three classes against four for the dense masks. 

The baseline shows that meaningful representations are learnt by a popular 
visual semantic segmentation architecture. These models succeed in detecting and classifying shapes of moving objects in raw radar representations even with sparse-point annotations. 
Performances on range-angle are not as good as in range-Doppler because the angular resolution of the sensor is low, resulting in less precise generated annotations. An extension to improve performances on this representation could be to transform it into Cartesian coordinates as done in \cite{major_vehicle_2019}.
For both representations, results are promising since the temporal dimension of the objects signatures has not yet been taken into account. 

\section{Discussions}
\label{discussions}

The semi-automatic algorithm presented in Section \ref{annotations} generates precise annotations on raw radar data, but it has limitations. Occlusion phenomena are problematic for tracking, since they lead to a disappearance of the object point cloud in the DoA-Doppler representation. An improvement could be to detect such occlusions in the video frames and include them in the tracking pipeline. The clustering in the DoA-Doppler representation is also a difficult task in specific cases. When objects are close to each other with a similar radial velocity, point clouds are difficult to distinguish. Further work on the bandwidth selection and optimisation of this selection could be explored.

The CARRADA dataset provides precise annotations to explore a range of supervised learning tasks. 
We propose a simple baseline for semantic segmentation trained on dense mask annotations. It could be extended by using  temporal information or both dense mask and sparse points annotations at the same time during training. Current architectures and loss functions could also be optimized for semantic segmentation of sparse ground-truth points.
Object detection could be considered by using bounding boxes to detect and classify object signatures. 
As off-the-shelf object detection algorithms are not adapted to the radar data representation and to the unusual size of the provided annotations, further work is required to redesign these methods.
%
By identifying and tracking specific instances of objects, other opportunities are opened.
Tracking of sparse points or bounding boxes could also be considered. 

\section{Conclusion}
\label{conclusion}

The CARRADA dataset contains synchronised video frames, range-angle and range-Doppler raw radar representations. Radar data are annotated with sparse points, bounding boxes and dense masks to localize and categorize the object signatures. A unique identification number is also provided for each instance. Annotations are generated using a semi-supervised algorithm based on visual and physical knowledge. The pipeline could be used to annotate any camera-radar recordings with similar settings. The dataset, code for the annotation algorithm and code for dataset visualisation are publicly available. 
We hope that this work will encourage other teams to record and release annotated radar datasets combined with other sensors.
This work also aims to motivate deep learning research applied to radar sensor and multi-sensor fusion for scene understanding.


\section*{Acknowledgment}
The authors would like to express their thanks to the Sensor Cortex team, which has recorded these data and spent time to answer our questions, and to Gabriel de Marmiesse for his valuable technical help.


\bibliographystyle{IEEEtranS}
\bibliography{icpr_biblio_origine}

\begin{thebibliography}{10}
\providecommand{\url}[1]{#1}
\csname url@samestyle\endcsname
\providecommand{\newblock}{\relax}
\providecommand{\bibinfo}[2]{#2}
\providecommand{\BIBentrySTDinterwordspacing}{\spaceskip=0pt\relax}
\providecommand{\BIBentryALTinterwordstretchfactor}{4}
\providecommand{\BIBentryALTinterwordspacing}{\spaceskip=\fontdimen2\font plus
\BIBentryALTinterwordstretchfactor\fontdimen3\font minus
  \fontdimen4\font\relax}
\providecommand{\BIBforeignlanguage}[2]{{%
\expandafter\ifx\csname l@#1\endcsname\relax
\typeout{** WARNING: IEEEtranS.bst: No hyphenation pattern has been}%
\typeout{** loaded for the language `#1'. Using the pattern for}%
\typeout{** the default language instead.}%
\else
\language=\csname l@#1\endcsname
\fi
#2}}
\providecommand{\BIBdecl}{\relax}
\BIBdecl

\bibitem{aldera_what_2019}
\BIBentryALTinterwordspacing
R.~Aldera, D.~D. Martini, M.~Gadd, and P.~Newman, ``What could go wrong?
  {I}ntrospective radar odometry in challenging environments,'' in \emph{ITSC},
  2019.
\BIBentrySTDinterwordspacing

\bibitem{barnes_oxford_2020}
\BIBentryALTinterwordspacing
D.~Barnes, M.~Gadd, P.~Murcutt, P.~Newman, and I.~Posner, ``The {Oxford}
  {Radar} {RobotCar} {dataset}: {A} {radar} {extension} to the {Oxford}
  {RobotCar} {dataset},'' in \emph{{ICRA}}, 2020.
\BIBentrySTDinterwordspacing

\bibitem{bewley_simple_2016}
\BIBentryALTinterwordspacing
A.~Bewley, Z.~Ge, L.~Ott, F.~Ramos, and B.~Upcroft, ``Simple {online} and
  {realtime} {tracking},'' in \emph{ICIP}, 2016.
\BIBentrySTDinterwordspacing

\bibitem{bijelic_benchmark_2018}
\BIBentryALTinterwordspacing
M.~Bijelic, T.~Gruber, and W.~Ritter, ``A {benchmark} for {lidar} {sensors} in
  {fog}: {Is} {detection} {breaking} {down}?'' in \emph{IV}, 2018.
\BIBentrySTDinterwordspacing

\bibitem{brooker_understanding_2005}
G.~Brooker, ``Understanding millimetre wave {FMCW} radars,'' in \emph{ICST},
  2005.

\bibitem{brooks_temporal_2018}
\BIBentryALTinterwordspacing
D.~A. Brooks, O.~Schwander, F.~Barbaresco, J.-Y. Schneider, and M.~Cord,
  ``Temporal {deep} {learning} for {drone} {micro}-{Doppler}
  {classification},'' in \emph{IRS}, 2018.
\BIBentrySTDinterwordspacing

\bibitem{bugeau_bandwidth_2007}
\BIBentryALTinterwordspacing
A.~Bugeau and P.~Pérez, ``Bandwidth selection for kernel estimation in mixed
  multi-dimensional spaces,'' INRIA, Tech. Rep. RR-6286, 2007.
\BIBentrySTDinterwordspacing

\bibitem{caesar_nuscenes_2020}
\BIBentryALTinterwordspacing
H.~Caesar, V.~Bankiti, A.~H. Lang, S.~Vora, V.~E. Liong, Q.~Xu, A.~Krishnan,
  Y.~Pan, G.~Baldan, and O.~Beijbom, ``{nuScenes}: {A} {multimodal} {dataset}
  for {autonomous} {driving},'' in \emph{{CVPR}}, 2020.
\BIBentrySTDinterwordspacing

\bibitem{capobianco_vehicle_2018}
\BIBentryALTinterwordspacing
S.~Capobianco, L.~Facheris, F.~Cuccoli, and S.~Marinai, ``Vehicle
  {classification} {based} on {convolutional} {networks} {applied} to {FMCW}
  {radar} {signals},'' \emph{IEEE {TRAP}}, vol. 728, 2018.
\BIBentrySTDinterwordspacing

\bibitem{comaniciu_mean_2002}
\BIBentryALTinterwordspacing
D.~Comaniciu and P.~Meer, ``Mean shift: A robust approach toward feature space
  analysis,'' \emph{IEEE T-PAMI}, vol.~24, no.~5, 2002.
\BIBentrySTDinterwordspacing

\bibitem{cordts_cityscapes_2016}
\BIBentryALTinterwordspacing
M.~Cordts, M.~Omran, S.~Ramos, T.~Rehfeld, M.~Enzweiler, R.~Benenson,
  U.~Franke, S.~Roth, and B.~Schiele, ``The {Cityscapes} {dataset} for
  {semantic} {urban} {scene} {understanding},'' in \emph{CVPR}, 2016.
\BIBentrySTDinterwordspacing

\bibitem{dekker_gesture_2017}
\BIBentryALTinterwordspacing
B.~Dekker, S.~Jacobs, A.~Kossen, M.~Kruithof, A.~Huizing, and M.~Geurts,
  ``Gesture recognition with a low power {FMCW} radar and a deep convolutional
  neural network,'' in \emph{EURAD}, 2017.
\BIBentrySTDinterwordspacing

\bibitem{endres_new_2003}
\BIBentryALTinterwordspacing
D.~Endres and J.~Schindelin, ``\BIBforeignlanguage{en}{A new metric for
  probability distributions},'' \emph{\BIBforeignlanguage{en}{IEEE T-IT}},
  vol.~49, no.~7, 2003.
\BIBentrySTDinterwordspacing

\bibitem{gao_experiments_2019}
\BIBentryALTinterwordspacing
X.~Gao, G.~Xing, S.~Roy, and H.~Liu, ``Experiments with {mmWave} {automotive}
  {radar} {test}-bed,'' in \emph{IEEE {ACSSC}}, 2019.
\BIBentrySTDinterwordspacing

\bibitem{geiger_vision_2013}
\BIBentryALTinterwordspacing
A.~Geiger, P.~Lenz, C.~Stiller, and R.~Urtasun,
  ``\BIBforeignlanguage{en}{Vision meets robotics: {The} {KITTI} dataset},''
  \emph{\BIBforeignlanguage{en}{JRR}}, vol.~32, no.~11, 2013.
\BIBentrySTDinterwordspacing

\bibitem{ghaleb_micro-doppler_2009}
A.~Ghaleb, ``\BIBforeignlanguage{French}{Micro-{Doppler} analysis of
  non-stationary moving targets in radar imaging},'' Ph.D. dissertation,
  Telecom Paris, 2009.

\bibitem{he_mask_2017}
\BIBentryALTinterwordspacing
K.~He, G.~Gkioxari, P.~Dollar, and R.~Girshick, ``Mask {R}-{CNN},'' in
  \emph{ICCV}, 2017.
\BIBentrySTDinterwordspacing

\bibitem{huang_apolloscape_2020}
\BIBentryALTinterwordspacing
X.~Huang, P.~Wang, X.~Cheng, D.~Zhou, Q.~Geng, and R.~Yang, ``The {ApolloScape}
  {open} {dataset} for {autonomous} {driving} and its {application},''
  \emph{IEEE T-PAMI}, vol.~42, no.~10, 2020.
\BIBentrySTDinterwordspacing

\bibitem{kim_human_2016}
\BIBentryALTinterwordspacing
Y.~Kim and T.~Moon, ``Human {detection} and {activity} {classification} {based}
  on {micro}-{Doppler} {signatures} {using} {deep} {convolutional} {neural}
  {networks},'' \emph{IEEE GRSL}, vol.~13, no.~1, 2016.
\BIBentrySTDinterwordspacing

\bibitem{kim_hand_2016}
\BIBentryALTinterwordspacing
Y.~Kim and B.~Toomajian, ``Hand {gesture} {recognition} {using}
  {micro}-{Doppler} {signatures} {with} {convolutional} {neural} {network},''
  \emph{IEEE Access}, vol.~4, pp. 7125--7130, 2016.
\BIBentrySTDinterwordspacing

\bibitem{kingma_adam_2015}
\BIBentryALTinterwordspacing
D.~P. Kingma and J.~Ba, ``Adam: {A} {method} for {stochastic} {optimization},''
  in \emph{ICLR}, 2015.
\BIBentrySTDinterwordspacing

\bibitem{klarenbeek_multi-target_2017}
\BIBentryALTinterwordspacing
G.~Klarenbeek, R.~I.~A. Harmanny, and L.~Cifola, ``Multi-target human gait
  classification using {LSTM} recurrent neural networks applied to
  micro-{Doppler},'' in \emph{EURAD}, 2017.
\BIBentrySTDinterwordspacing

\bibitem{kullback_information_1951}
\BIBentryALTinterwordspacing
S.~Kullback and R.~A. Leibler, ``\BIBforeignlanguage{en}{On {information} and
  {sufficiency}},'' \emph{\BIBforeignlanguage{en}{The Annals of Mathematical
  Statistics}}, vol.~22, no.~1, 1951.
\BIBentrySTDinterwordspacing

\bibitem{lekic_automotive_2019}
\BIBentryALTinterwordspacing
V.~Lekic and Z.~Babic, ``\BIBforeignlanguage{en}{Automotive radar and camera
  fusion using {generative} {adversarial} {networks}},''
  \emph{\BIBforeignlanguage{en}{CVIU}}, vol. 184, pp. 1--8, 2019.
\BIBentrySTDinterwordspacing

\bibitem{lim_radar_2019}
T.-Y. Lim, A.~Ansari, B.~Major, D.~Fontijne, M.~Hamilton, R.~Gowaikar, and
  S.~Subramanian, ``Radar and {camera} {early} {fusion} for {vehicle}
  {detection} in {advanced} {driver} {assistance} {systems},'' in
  \emph{NeurIPS}, 2019.

\bibitem{long_fully_2015}
\BIBentryALTinterwordspacing
J.~Long, E.~Shelhamer, and T.~Darrell, ``Fully convolutional networks for
  semantic segmentation,'' in \emph{CVPR}, 2015.
\BIBentrySTDinterwordspacing

\bibitem{major_vehicle_2019}
\BIBentryALTinterwordspacing
B.~Major, D.~Fontijne, A.~Ansari, R.~T. Sukhavasi, R.~Gowaikar, M.~Hamilton,
  S.~Lee, S.~Grzechnik, and S.~Subramanian, ``Vehicle {detection} {with}
  {automotive} {radar} {using} {deep} {learning} on {range}-{azimuth}-{doppler}
  {tensors},'' in \emph{ICCV Workshop}, 2019.
\BIBentrySTDinterwordspacing

\bibitem{meyer_automotive_2019}
M.~Meyer and G.~Kuschk, ``Automotive {radar} {dataset} for {deep} {learning}
  {based} {3D} {object} {detection},'' in \emph{EuRAD}, 2019.

\bibitem{molchanov_multi-sensor_2015}
\BIBentryALTinterwordspacing
P.~Molchanov, S.~Gupta, K.~Kim, and K.~Pulli, ``Multi-sensor system for
  driver's hand-gesture recognition,'' in \emph{FG}, 2015.
\BIBentrySTDinterwordspacing

\bibitem{nabati_rrpn_2019}
\BIBentryALTinterwordspacing
R.~Nabati and H.~Qi, ``{RRPN}: {Radar} {region} {proposal} {network} for
  {object} {detection} in {autonomous} {vehicles},'' in \emph{ICIP}, 2019.
\BIBentrySTDinterwordspacing

\bibitem{nowruzi_deep_2020}
F.~E. Nowruzi, D.~Kolhatkar, P.~Kapoor, E.~J. Heravi, R.~Laganiere, J.~Rebut,
  and W.~Malik, ``Deep open space segmentation using automotive radar,'' in
  \emph{ICMIM}, 2020.

\bibitem{patel_deep_2019}
\BIBentryALTinterwordspacing
K.~Patel, K.~Rambach, T.~Visentin, D.~Rusev, M.~Pfeiffer, and B.~Yang, ``Deep
  {learning}-based {object} {classification} on {automotive} {radar}
  {spectra},'' in \emph{RadarConf}, 2019.
\BIBentrySTDinterwordspacing

\bibitem{rohling_radar_1983}
\BIBentryALTinterwordspacing
H.~Rohling, ``Radar {CFAR} {thresholding} in {clutter} and {multiple} {target}
  {situations},'' \emph{IEEE T-AES}, vol. AES-19, no.~4, 1983.
\BIBentrySTDinterwordspacing

\bibitem{sun_scalability_2020}
\BIBentryALTinterwordspacing
P.~Sun, H.~Kretzschmar, X.~Dotiwalla, A.~Chouard, V.~Patnaik, P.~Tsui, J.~Guo,
  Y.~Zhou, Y.~Chai, B.~Caine, V.~Vasudevan, W.~Han, J.~Ngiam, H.~Zhao,
  A.~Timofeev, S.~Ettinger, M.~Krivokon, A.~Gao, A.~Joshi, Y.~Zhang, J.~Shlens,
  Z.~Chen, and D.~Anguelov, ``Scalability in {perception} for {autonomous}
  {driving}: {Waymo} {Open} {dataset},'' in \emph{{CVPR}}, 2020.
\BIBentrySTDinterwordspacing

\bibitem{sun_automatic_2019}
\BIBentryALTinterwordspacing
Y.~Sun, T.~Fei, S.~Gao, and N.~Pohl, ``Automatic {radar}-based {gesture}
  {detection} and {classification} via a {region}-based {deep} {convolutional}
  {neural} {network},'' in \emph{ICASPP}, 2019.
\BIBentrySTDinterwordspacing

\bibitem{wang_interacting_2016}
\BIBentryALTinterwordspacing
S.~Wang, J.~Song, J.~Lien, I.~Poupyrev, and O.~Hilliges,
  ``\BIBforeignlanguage{en}{Interacting with {Soli}: {Exploring}
  {fine}-{grained} {dynamic} {gesture} {recognition} in the {radio}-{frequency}
  {spectrum}},'' in \emph{\BIBforeignlanguage{en}{UIST}}, 2016.
\BIBentrySTDinterwordspacing

\bibitem{wang_rammar_2019}
\BIBentryALTinterwordspacing
Y.~Wang, X.~Jia, M.~Zhou, X.~Yang, and Z.~Tian, ``Rammar: {RAM} {assisted}
  {mask} {R}-{CNN} for {FMCW} {sensor} {based} {HGD} {system},'' in \emph{ICC},
  2019.
\BIBentrySTDinterwordspacing

\bibitem{weston_probably_2019}
\BIBentryALTinterwordspacing
R.~Weston, S.~Cen, P.~Newman, and I.~Posner, ``Probably {unknown}: {Deep}
  {inverse} {sensor} {modelling} {radar},'' in \emph{{ICRA}}, 2019.
\BIBentrySTDinterwordspacing

\bibitem{yu_bdd100k_2020}
\BIBentryALTinterwordspacing
F.~Yu, W.~Xian, Y.~Chen, F.~Liu, M.~Liao, V.~Madhavan, and T.~Darrell,
  ``{BDD100K}: {A} {diverse} {driving} {video} {database} with {scalable}
  {annotation} {tooling},'' \emph{{CVPR}}, 2020.
\BIBentrySTDinterwordspacing

\bibitem{zhang_object_2020}
\BIBentryALTinterwordspacing
G.~Zhang, H.~Li, and F.~Wenger, ``Object {detection} and 3d {estimation} {via}
  an {FMCW} {radar} {using} a {fully} {convolutional} {network},'' in
  \emph{{ICASSP}}, 2020.
\BIBentrySTDinterwordspacing

\bibitem{zhang_u-deephand_2019}
\BIBentryALTinterwordspacing
Z.~Zhang, Z.~Tian, Y.~Zhang, M.~Zhou, and B.~Wang, ``u-{DeepHand}: {FMCW}
  {radar}-{based} {unsupervised} {hand} {gesture} {feature} {learning} {using}
  {deep} {convolutional} {auto}-{encoder} {network},'' \emph{IEEE Sensors
  Journal}, vol.~19, no.~16, 2019.
\BIBentrySTDinterwordspacing

\bibitem{zhang_latern_2018}
\BIBentryALTinterwordspacing
Z.~Zhang, Z.~Tian, and M.~Zhou, ``Latern: {Dynamic} {continuous} {hand}
  {gesture} {recognition} {using} {FMCW} {radar} {sensor},'' \emph{IEEE Sensors
  Journal}, vol.~18, no.~8, 2018.
\BIBentrySTDinterwordspacing

\end{thebibliography}

\end{document}